\documentclass{article}

\usepackage{PRIMEarxiv}

\usepackage[utf8]{inputenc} 
\usepackage[T1]{fontenc}    
\usepackage{hyperref}       
\usepackage{url}            
\usepackage{booktabs}       
\usepackage{amsmath}        
\usepackage{amsfonts}       
\usepackage{nicefrac}       
\usepackage{microtype}      
\usepackage{lipsum}
\usepackage{fancyhdr}       
\DeclareUnicodeCharacter{2248}{\ensuremath{\approx}}
\usepackage{graphicx}       
\usepackage{enumitem}
\graphicspath{{media/}}     

\title{Alignment is Localized: A Causal Probe into Preference Layers
}

\author{
  Archie Chaudhury\\
  Independent \\
   \\
  Atlanta, GA\\
  \texttt{archchaudhury02@gmail.com} \\
}

\begin{document}
\maketitle

\begin{abstract}
Reinforcement Learning frameworks, particularly those utilizing human annotations, have become an increasingly popular method for preference fine-tuning, where the outputs of a language model are tuned to match a certain set of behavioral policies or guidelines. Reinforcement Learning through Human Feedback (RLHF) is perhaps the most popular implementation of such a framework, particularly for aligning LMs toward safety and human intent. However, the internal workings of how such alignment is achieved remain largely opaque. In this work, we systematically analyze preference optimization for language model alignment by applying layer-wide causal patching between a base model and its tuned counterpart across human preference pairs. We implement our methodology on \textit{Llama-3.2-1B}, and find that alignment is spatially localized: mid-layer activations encode a distinct subspace that causally determines reward-consistent behavior, while early and late layers remain largely unaffected. Utilizing LASSO regression, we also find that only a small number of layers possess non-zero coefficients linking activation distances to reward gains. Overall, we show that, at least for some language models, alignment from human-based, preferential tuning is a directional, low rank process, rather than diffuse and parameteric. 
\end{abstract}


\section{Introduction}
Aligning artificial intelligence systems, such as Language Models (LMs), is perhaps the most important cornerstone in contemporary artificial intelligence research. Modern frontier AI labs have seem to convened on utilizing human feedback combined with reinforcement learning to tune their models, assuming that the models they create, on their own, are misaligned by default to some degree. Techniques such as Reinforcement Learning from Human Feedback (RLHF) \cite{christiano2023deepreinforcementlearninghuman} and Direct Preference Optimization (DPO) \cite{rafailov2024directpreferenceoptimizationlanguage} have enabled language models to generate outputs that are more helpful, honest, and harmless. These methods work by tuning an existing model toward behavior that is preferred by humans through labeled datasets, turning generalized token prediction systems into agents that can maintain a conversation with a human user and take actions on their behalf \cite{park2023generativeagentsinteractivesimulacra}.

Yet, despite the sustained success of such methods, the internal mechanisms behind how a model reaches a degree of alignment with a human-dictated policy remains poorly understood. It is still unclear whether alignment is a diffuse phenomenon that affects all of the layers of a model, or whether it is localized, mediated through specific internal subspace that causally influence model behavior.

In this work, we propose that preference-based alignment is indeed low-dimensional, and operates casually, influencing model behavior through a small set of internal directions that allow it to maintain its baseline knowledge, rather than a broad change. To empirically investigate this hypothesis, we conduct a series of causal intervention experiments between a base and preference-tuned variant of Llama-3.2-1B.
Using activation patching \cite{meng2023locatingeditingfactualassociations}, we transplant hidden activations between models layer by layer and measure the resulting change in reward-aligned log-likelihoods. We further employ LASSO regression to identify which layers’ representational distances best predict reward improvements.
The resulting sparse coefficients coincide with the mid-layer causal bottleneck, providing convergent evidence that alignment is both sparse and localized.

\begin{figure}[h]
  \centering
  \includegraphics[width=0.60\linewidth]{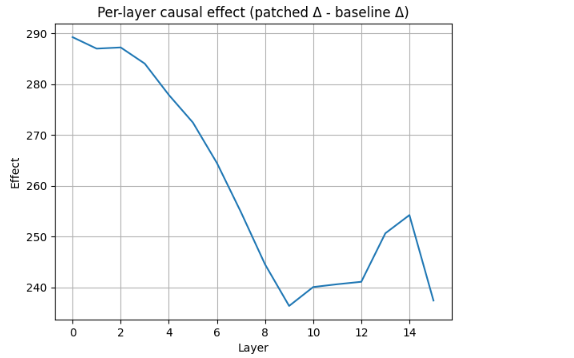}
  \vspace{-0.5em}
  \caption{
  Mean per-layer \emph{causal effect} of activation patching (DPO$\to$Base) on the Anthropic HHH preference pairs.}
  \label{fig:intro-teaser}
  \vspace{-0.5em}
\end{figure}

Our primary findings utilizing a standard RLHF dataset are as follows: Patching activations from the aligned model into the base model significantly increased preference-consistent behavior. Alignment effects are concentrated in a narrow range of mid-stack layers, with early and late layers showing little causal contribution. A small set of principal activation directions reproduce most of the alignment effect. 

Our experiments, although only done on one model pair, suggest that the same causal pathways that govern token prediction also mediate reinforcement learning gradients focused on alignment.

\section{Related Work}
Reinforcement Learning from Human Feedback (RLHF) has emerged as the dominant paradigm for aligning large language models with human preferences \cite{ouyang2022traininglanguagemodelsfollow}.
Direct Preference Optimization (DPO) reformulates preference fine-tuning as a closed-form policy optimization problem in logit space.

Mechanistic interpretability seeks to reverse-engineer neural networks into human-understandable components, identifying how specific neurons, heads, or circuits implement high-level functions \cite{nanda2023progressmeasuresgrokkingmechanistic}.
Recent work has mapped interpretable features such as factual recall, induction heads \cite{elhage2021mathematical}, and algorithmic reasoning. Activation patching has become a powerful tool in mechanistic interpretability, enabling researchers to isolate the causal contributions of specific activations or layers to model behavior.
These efforts demonstrate that high-level behaviors can often be localized to small, linearly separable subspaces within model representations, particularly Anthropic's analysis of RLHF models, which suggested that distinct subspaces may encode helpfulness and harmlessness preferences \cite{bai2022traininghelpfulharmlessassistant}.

Numerous studies have demonstrated that fine-tuning and transfer learning often operate in low-dimensional subspaces of the full parameter or activation space.
Methods such as Low-Rank Adaptation (LoRA) \cite{hu2021loralowrankadaptationlarge} show that large-scale models can be specialized using only a small number of learned directions.
\section{Methodology}
Our primary goal was to determine where and how reinforcment learning aimed at producing alignment was encoded within Large Language Models (LLMs). We conduct all experiments using the Llama 3.2 1B model family released by Meta AI (2025) \cite{grattafiori2024llama3herdmodels}.
Specifically, we compare the base checkpoint (Llama-3.2-1B) with its instruction-tuned variant (Llama-3.2-1B-Instruct), which has been fine-tuned via a supervised finetuning process.
Both checkpoints are publicly available on Hugging Face, share identical tokenizers, and maintain the same model architecture, ensuring feature-space compatibility for causal interventions.

For preference supervision, we use the Anthropic Helpful–Harmless–Honest (HHH) dataset \cite{bai2022traininghelpfulharmlessassistant}, which contains human-annotated pairs of responses labeled as chosen or rejected according to safety and helpfulness criteria.
From this corpus, we sample 80 pairs spanning a mixture of helpfulness and harmlessness scenarios to ensure diversity in alignment-relevant behaviors. Each prompt pair is a distinct test-case that allows us to measure how human preference manifests itself in internal representations.

\subsection{Activation Patching}

For a given prompt, with a positive completion $y^{+}$ and a negative completion $y^{-}$, we implement activation patching by first measuring the overall difference in the conditional log probablities of the response: 
\begin{equation}
\Delta \log p = \log p(y^{+} \mid x) - \log p(y^{-} \mid x),
\end{equation}

A greater $\Delta \log p$ value indicates that the response was more closely aligned with the human annotations corresponding to the postive, aligned responses. 

We record hidden activations from both the base and tuned models for each transformer layer during forward passes on the same input sequences.
In the \textit{activation patching} step, we replace a single layer’s hidden states in one model (the \textit{target}) with those from the other model (the \textit{source}), keeping all other layers fixed.
We then recompute the model’s output log-probabilities and evaluate how this intervention modifies $\Delta \log p$.
This difference reflects the \textbf{causal contribution} of that layer’s activations to alignment behavior. The effect of the tuned model on the baseline model is shown in Figure~\ref{fig:intro-teaser}.

\subsection{Linear Probes and Sparse Regression Attribution}

To complement the causal patching experiments, we use linear probing and sparse regression to quantify where alignment information is represented within the network. 
For each preference pair, we compute the per-layer activation difference between the tuned and base models:
\begin{equation}
\Delta h_\ell = h_\ell^{(\text{aligned})} - h_\ell^{(\text{base})},
\end{equation}
where $h_\ell$ denotes the mean hidden representation of the final token at layer~$\ell$.
We then fit a simple linear probe to predict the behavioral alignment signal (the log-probability margin between preferred and rejected completions):
\begin{equation}
\hat{y}_\ell = w_\ell^\top \|\Delta h_\ell\|_2 + b_\ell,
\end{equation}
providing an interpretable measure of how each layer’s representational shift correlates with aligned behavior. 

To identify the minimal set of influential layers, we apply LASSO regression~\cite{10.1111/j.2517-6161.1996.tb02080.x}:
\begin{equation}
\min_{w} \ \|Y - Xw\|_2^2 + \lambda \|w\|_1,
\end{equation}
where $X$ contains per-layer activation distances $\|\Delta h_\ell\|_2$ and $Y$ is the observed causal effect ($\Delta \log p$) from patching.
The resulting sparse coefficients highlight only the layers most predictive of alignment gains.

\subsection{Complementary Experiments}
We also performed a set of complemntary experiements, including abalations and tuning to ensure our results were not one off, to further confirm our core hypothesis. Some of these experiments, such as the low rank reconstruction, actually provided new insights. 

\begin{description}[leftmargin=1.5em,labelindent=0em]
    \item[(A1) Directionality.] 
    Patching activations from the tuned model into the base consistently increases $\Delta \log p$. We reverse this to see if it has any effect. A full set of results can found in section \ref{appendix:c} of the appendix. 

    \item[(A2) Random and Identity Controls.]
    To verify that observed effects are not spurious, we patch randomized activations and self-replacement controls. A full set of results can be found in section \ref{appendix:a} of the appendix. 

    \item[(A3) $\boldsymbol{\alpha}$-Interpolation (Dose--Response).]
    We scale the strength of the causal intervention with a mixing coefficient $\alpha \in [0, 1]$.
    The results can be found in section \ref{appendix:b} of the appendix. 

    \item[(A4) Source Variants.]
    We patch activations corresponding to the \emph{chosen}, \emph{rejected}, and (\emph{chosen}--\emph{rejected}) differences separately, seeing if this changes the alignment signal. 

    \item[(A5) Low-Rank Reconstruction.]
    We perform singular value decomposition (SVD) on the tuned activations and patch only the top-$k$ principal components. 
\end{description}

\section{Results}
\label{sec:Results}
Our results show that human preference alignment emerges from a small, localized subspace of mid-layer representations. Specifically, for the small subset of dataset that we used and our model pair, layer 8 was where the majority of the reward associated with positive responses was concentrated. 

\subsection{Sparse Attribution (LASSO Coefficients)}

\begin{figure}[h!]
\centering
\includegraphics[width=0.55\linewidth]{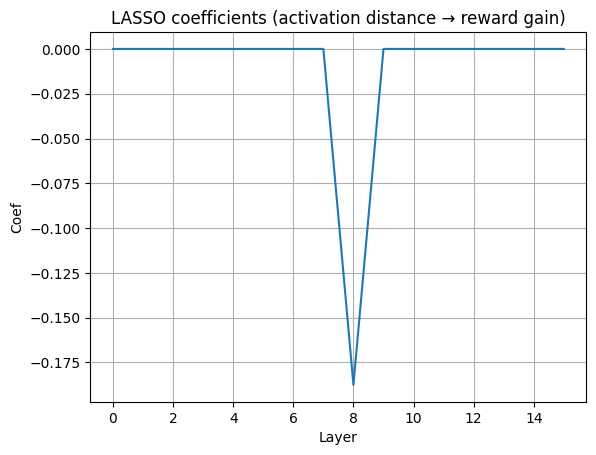}
\caption{LASSO coefficients linking activation distance to reward gain ($\Delta \log p$). A single mid-layer dominates the attribution.}
\label{fig:lasso}
\end{figure}

Table~\ref{tab:lasso} reports the per-layer coefficients from the LASSO regression linking activation distance to reward gain (see Fig.~\ref{fig:lasso}).  
Only layer~8 exhibits a significant non-zero coefficient, confirming a sparse and highly localized attribution pattern.

\begin{table}[h!]
\centering
\caption{Per-layer LASSO coefficients predicting reward gain ($\Delta \log p$) from activation distances.}
\label{tab:lasso}
\begin{tabular}{c|rrrrrrrrrrrrrrrr}
\toprule
Layer & 0 & 1 & 2 & 3 & 4 & 5 & 6 & 7 & 8 & 9 & 10 & 11 & 12 & 13 & 14 & 15 \\ \midrule
Coeff & 0.00 & 0.00 & 0.00 & 0.00 & 0.00 & 0.00 & 0.00 & 0.00 & \textbf{-0.18} & 0.00 & 0.00 & 0.00 & 0.00 & 0.00 & 0.00 & 0.00 \\
\bottomrule
\end{tabular}
\end{table}

The resulting sparse coefficient pattern mirrors the causal patching peaks observed in the main experiment, reinforcing that alignment mechanisms are concentrated within a single mid-layer bottleneck.

\subsection{Source Variant Causality}

\begin{figure}[h!]
\centering
\includegraphics[width=0.55\linewidth]{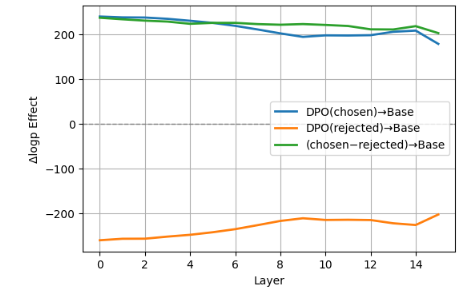}
\caption{Causal patching results for source variants. The contrastive $(y^+ - y^-)$ representation drives positive alignment effects, while rejected activations reduce $\Delta \log p$.}
\label{fig:source_variant}
\end{figure}

Figure~\ref{fig:source_variant} compares causal effects for activations derived from the chosen, rejected, and contrastive (chosen--rejected) representations. These representation can be defined as follows. The chosen representation represents the tuned model's internal activations when producing the aligned (human-preferred) response. In turn, the rejected representation is the tuned model's internal trajectory as it produces a dispreferred response. The contrastive representation is the difference vector between the two: the direction in activation space that separates aligned from misaligned behavior.
The chosen and contrastive variants yield positive $\Delta \log p$ effects across layers, while the rejected variant produces the opposite trend.

This pattern supports the hypothesis that alignment representations encode a directed reward signal, rather than being a generalized feature transfer. 

\subsection{Low-Rank Reconstruction}

\begin{figure}[h!]
\centering
\includegraphics[width=0.5\linewidth]{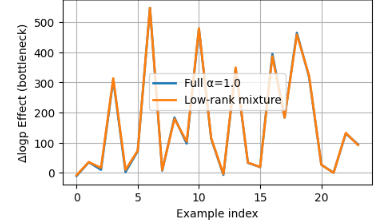}
\caption{Low-rank reconstruction results. A small number of principal components capture nearly all causal alignment effect.}
\label{fig:c9_svd}
\end{figure}

To test whether alignment lies in a low-dimensional subspace, making it low rank in addition to being localized, we perform singular value decomposition (SVD) on tuned activations and patch only the top-$k$ components. We do this by collecting tuned activations $H_\ell \in \mathbb{R}^{N \times d}$ 
from layer~$\ell$ across all tokens and examples:
\[
H_\ell = U \Sigma V^{\top}.
\]
We then reconstruct a low-rank approximation using only the top-$k$ singular components:
\begin{equation}
H_{\ell, \text{approx}} = U_{[:, :k]} \Sigma_{[:k, :k]} V_{[:, :k]}^{\top}.
\end{equation}
This approximation preserves the dominant modes of variance within the aligned representation. 
We patch $H_{\ell, \text{approx}}$ into the base model in place of the full tuned activations 
and remeasure the causal alignment effect ($\Delta \log p$).
As shown in Figure \ref{fig:c9_svd}, retaining just 4 components (≈15\% of total variance) reproduces nearly the full alignment effect.
\begin{table}[h!]
\centering
\caption{Low-rank reconstruction summary for (layer 9).}
\label{tab:svd}
\begin{tabular}{lccc}
\toprule
Rank ($k$) & Approx.~Mean & Full~Mean & Variance Ratio \\
\midrule
4 & 161.6 & 160.3 & 0.986 \\
\bottomrule
\end{tabular}
\end{table}

\section{Discussion}
Our experiments, although limited in scope, provide evidence that preferential, human annotated alignment is localized, directional, and low-rank. Across all causal interventions, patching midlayer activations from the preference-aligned Llama 3.2 Instruct model into its corresponding base model consistently increased the occurrence of aligned behavior. Linear probes and sparse regression (LASSO) revealed that alignment information is not uniformly distributed across the network but concentrated in a small set of middle layers. SVD analysis revealed that patching only the top proportion of activations could produce the full effect of alignment, indicating that alignment resides in a compact, low-dimensional subspace.

\subsection{Limitations and Future Work}
As previously stated, our experiment was limited, focusing only only on a single model family and a subset of a single dataset. Our results may differ for models with different architectures, or for datasets with more diverse or different prompts. 

Our experiment also focused exclusively on linear, layer-based interventions. Studying non-linear, or cross-layer interventions, could be an interesting direction for future work. Future work could also focus on testing the transferability of discovered subspaces across models, modalities, and alignment dimensions, such as honesty, or less explicit behavioral patterns such as sycophancy.

\section{Conclusion}
As the development of LMs continues, ensuring that they are aligned and safe will continue to be priority. This will be driven in large part by human feedback that dictates the manner in which models behave in production settings. Our experiment was concentrated on uncovering where and how human feedback is utilized within a model to make it more aligned. 

By combining activation patching, linear probing, and low-rank reconstruction, we show that preference alignment is directional, sparse, and ultimately localized within a mid-layer bottleneck.

\bibliographystyle{unsrt}  
\bibliography{references}  

\begin{thebibliography}{10}

\bibitem{christiano2023deepreinforcementlearninghuman}
Paul Christiano, Jan Leike, Tom~B. Brown, Miljan Martic, Shane Legg, and Dario Amodei.
\newblock Deep reinforcement learning from human preferences, 2023.

\bibitem{rafailov2024directpreferenceoptimizationlanguage}
Rafael Rafailov, Archit Sharma, Eric Mitchell, Stefano Ermon, Christopher~D. Manning, and Chelsea Finn.
\newblock Direct preference optimization: Your language model is secretly a reward model, 2024.

\bibitem{park2023generativeagentsinteractivesimulacra}
Joon~Sung Park, Joseph~C. O'Brien, Carrie~J. Cai, Meredith~Ringel Morris, Percy Liang, and Michael~S. Bernstein.
\newblock Generative agents: Interactive simulacra of human behavior, 2023.

\bibitem{meng2023locatingeditingfactualassociations}
Kevin Meng, David Bau, Alex Andonian, and Yonatan Belinkov.
\newblock Locating and editing factual associations in gpt, 2023.

\bibitem{ouyang2022traininglanguagemodelsfollow}
Long Ouyang, Jeff Wu, Xu~Jiang, Diogo Almeida, Carroll~L. Wainwright, Pamela Mishkin, Chong Zhang, Sandhini Agarwal, Katarina Slama, Alex Ray, John Schulman, Jacob Hilton, Fraser Kelton, Luke Miller, Maddie Simens, Amanda Askell, Peter Welinder, Paul Christiano, Jan Leike, and Ryan Lowe.
\newblock Training language models to follow instructions with human feedback, 2022.

\bibitem{nanda2023progressmeasuresgrokkingmechanistic}
Neel Nanda, Lawrence Chan, Tom Lieberum, Jess Smith, and Jacob Steinhardt.
\newblock Progress measures for grokking via mechanistic interpretability, 2023.

\bibitem{elhage2021mathematical}
Nelson Elhage, Neel Nanda, Catherine Olsson, Tom Henighan, Nicholas Joseph, Ben Mann, Amanda Askell, Yuntao Bai, Anna Chen, Tom Conerly, Nova DasSarma, Dawn Drain, Deep Ganguli, Zac Hatfield-Dodds, Danny Hernandez, Andy Jones, Jackson Kernion, Liane Lovitt, Kamal Ndousse, Dario Amodei, Tom Brown, Jack Clark, Jared Kaplan, Sam McCandlish, and Chris Olah.
\newblock A mathematical framework for transformer circuits.
\newblock {\em Transformer Circuits Thread}, 2021.
\newblock https://transformer-circuits.pub/2021/framework/index.html.

\bibitem{bai2022traininghelpfulharmlessassistant}
Yuntao Bai, Andy Jones, Kamal Ndousse, Amanda Askell, Anna Chen, Nova DasSarma, Dawn Drain, Stanislav Fort, Deep Ganguli, Tom Henighan, Nicholas Joseph, Saurav Kadavath, Jackson Kernion, Tom Conerly, Sheer El-Showk, Nelson Elhage, Zac Hatfield-Dodds, Danny Hernandez, Tristan Hume, Scott Johnston, Shauna Kravec, Liane Lovitt, Neel Nanda, Catherine Olsson, Dario Amodei, Tom Brown, Jack Clark, Sam McCandlish, Chris Olah, Ben Mann, and Jared Kaplan.
\newblock Training a helpful and harmless assistant with reinforcement learning from human feedback, 2022.

\bibitem{hu2021loralowrankadaptationlarge}
Edward~J. Hu, Yelong Shen, Phillip Wallis, Zeyuan Allen-Zhu, Yuanzhi Li, Shean Wang, Lu~Wang, and Weizhu Chen.
\newblock Lora: Low-rank adaptation of large language models, 2021.

\bibitem{grattafiori2024llama3herdmodels}
Aaron Grattafiori, Abhimanyu Dubey, Abhinav Jauhri, Abhinav Pandey, Abhishek Kadian, Ahmad Al-Dahle, Aiesha Letman, Akhil Mathur, Alan Schelten, Alex Vaughan, Amy Yang, Angela Fan, Anirudh Goyal, Anthony Hartshorn, Aobo Yang, Archi Mitra, Archie Sravankumar, Artem Korenev, Arthur Hinsvark, Arun Rao, Aston Zhang, Aurelien Rodriguez, Austen Gregerson, Ava Spataru, Baptiste Roziere, Bethany Biron, Binh Tang, Bobbie Chern, Charlotte Caucheteux, Chaya Nayak, Chloe Bi, Chris Marra, Chris McConnell, Christian Keller, Christophe Touret, Chunyang Wu, Corinne Wong, Cristian~Canton Ferrer, Cyrus Nikolaidis, Damien Allonsius, Daniel Song, Danielle Pintz, Danny Livshits, Danny Wyatt, David Esiobu, Dhruv Choudhary, Dhruv Mahajan, Diego Garcia-Olano, Diego Perino, Dieuwke Hupkes, Egor Lakomkin, Ehab AlBadawy, Elina Lobanova, Emily Dinan, Eric~Michael Smith, Filip Radenovic, Francisco Guzmán, Frank Zhang, Gabriel Synnaeve, Gabrielle Lee, Georgia~Lewis Anderson, Govind Thattai, Graeme Nail, Gregoire Mialon, Guan Pang,
  Guillem Cucurell, Hailey Nguyen, Hannah Korevaar, Hu~Xu, Hugo Touvron, Iliyan Zarov, Imanol~Arrieta Ibarra, Isabel Kloumann, Ishan Misra, Ivan Evtimov, Jack Zhang, Jade Copet, Jaewon Lee, Jan Geffert, Jana Vranes, Jason Park, Jay Mahadeokar, Jeet Shah, Jelmer van~der Linde, Jennifer Billock, Jenny Hong, Jenya Lee, Jeremy Fu, Jianfeng Chi, Jianyu Huang, Jiawen Liu, Jie Wang, Jiecao Yu, Joanna Bitton, Joe Spisak, Jongsoo Park, Joseph Rocca, Joshua Johnstun, Joshua Saxe, Junteng Jia, Kalyan~Vasuden Alwala, Karthik Prasad, Kartikeya Upasani, Kate Plawiak, Ke~Li, Kenneth Heafield, Kevin Stone, Khalid El-Arini, Krithika Iyer, Kshitiz Malik, Kuenley Chiu, Kunal Bhalla, Kushal Lakhotia, Lauren Rantala-Yeary, Laurens van~der Maaten, Lawrence Chen, Liang Tan, Liz Jenkins, Louis Martin, Lovish Madaan, Lubo Malo, Lukas Blecher, Lukas Landzaat, Luke de~Oliveira, Madeline Muzzi, Mahesh Pasupuleti, Mannat Singh, Manohar Paluri, Marcin Kardas, Maria Tsimpoukelli, Mathew Oldham, Mathieu Rita, Maya Pavlova, Melanie Kambadur,
  Mike Lewis, Min Si, Mitesh~Kumar Singh, Mona Hassan, Naman Goyal, Narjes Torabi, Nikolay Bashlykov, Nikolay Bogoychev, Niladri Chatterji, Ning Zhang, Olivier Duchenne, Onur Çelebi, Patrick Alrassy, Pengchuan Zhang, Pengwei Li, Petar Vasic, Peter Weng, Prajjwal Bhargava, Pratik Dubal, Praveen Krishnan, Punit~Singh Koura, Puxin Xu, Qing He, Qingxiao Dong, Ragavan Srinivasan, Raj Ganapathy, Ramon Calderer, Ricardo~Silveira Cabral, Robert Stojnic, Roberta Raileanu, Rohan Maheswari, Rohit Girdhar, Rohit Patel, Romain Sauvestre, Ronnie Polidoro, Roshan Sumbaly, Ross Taylor, Ruan Silva, Rui Hou, Rui Wang, Saghar Hosseini, Sahana Chennabasappa, Sanjay Singh, Sean Bell, Seohyun~Sonia Kim, Sergey Edunov, Shaoliang Nie, Sharan Narang, Sharath Raparthy, Sheng Shen, Shengye Wan, Shruti Bhosale, Shun Zhang, Simon Vandenhende, Soumya Batra, Spencer Whitman, Sten Sootla, Stephane Collot, Suchin Gururangan, Sydney Borodinsky, Tamar Herman, Tara Fowler, Tarek Sheasha, Thomas Georgiou, Thomas Scialom, Tobias Speckbacher,
  Todor Mihaylov, Tong Xiao, Ujjwal Karn, Vedanuj Goswami, Vibhor Gupta, Vignesh Ramanathan, Viktor Kerkez, Vincent Gonguet, Virginie Do, Vish Vogeti, Vítor Albiero, Vladan Petrovic, Weiwei Chu, Wenhan Xiong, Wenyin Fu, Whitney Meers, Xavier Martinet, Xiaodong Wang, Xiaofang Wang, Xiaoqing~Ellen Tan, Xide Xia, Xinfeng Xie, Xuchao Jia, Xuewei Wang, Yaelle Goldschlag, Yashesh Gaur, Yasmine Babaei, Yi~Wen, Yiwen Song, Yuchen Zhang, Yue Li, Yuning Mao, Zacharie~Delpierre Coudert, Zheng Yan, Zhengxing Chen, Zoe Papakipos, Aaditya Singh, Aayushi Srivastava, Abha Jain, Adam Kelsey, Adam Shajnfeld, Adithya Gangidi, Adolfo Victoria, Ahuva Goldstand, Ajay Menon, Ajay Sharma, Alex Boesenberg, Alexei Baevski, Allie Feinstein, Amanda Kallet, Amit Sangani, Amos Teo, Anam Yunus, Andrei Lupu, Andres Alvarado, Andrew Caples, Andrew Gu, Andrew Ho, Andrew Poulton, Andrew Ryan, Ankit Ramchandani, Annie Dong, Annie Franco, Anuj Goyal, Aparajita Saraf, Arkabandhu Chowdhury, Ashley Gabriel, Ashwin Bharambe, Assaf Eisenman, Azadeh
  Yazdan, Beau James, Ben Maurer, Benjamin Leonhardi, Bernie Huang, Beth Loyd, Beto~De Paola, Bhargavi Paranjape, Bing Liu, Bo~Wu, Boyu Ni, Braden Hancock, Bram Wasti, Brandon Spence, Brani Stojkovic, Brian Gamido, Britt Montalvo, Carl Parker, Carly Burton, Catalina Mejia, Ce~Liu, Changhan Wang, Changkyu Kim, Chao Zhou, Chester Hu, Ching-Hsiang Chu, Chris Cai, Chris Tindal, Christoph Feichtenhofer, Cynthia Gao, Damon Civin, Dana Beaty, Daniel Kreymer, Daniel Li, David Adkins, David Xu, Davide Testuggine, Delia David, Devi Parikh, Diana Liskovich, Didem Foss, Dingkang Wang, Duc Le, Dustin Holland, Edward Dowling, Eissa Jamil, Elaine Montgomery, Eleonora Presani, Emily Hahn, Emily Wood, Eric-Tuan Le, Erik Brinkman, Esteban Arcaute, Evan Dunbar, Evan Smothers, Fei Sun, Felix Kreuk, Feng Tian, Filippos Kokkinos, Firat Ozgenel, Francesco Caggioni, Frank Kanayet, Frank Seide, Gabriela~Medina Florez, Gabriella Schwarz, Gada Badeer, Georgia Swee, Gil Halpern, Grant Herman, Grigory Sizov, Guangyi, Zhang, Guna
  Lakshminarayanan, Hakan Inan, Hamid Shojanazeri, Han Zou, Hannah Wang, Hanwen Zha, Haroun Habeeb, Harrison Rudolph, Helen Suk, Henry Aspegren, Hunter Goldman, Hongyuan Zhan, Ibrahim Damlaj, Igor Molybog, Igor Tufanov, Ilias Leontiadis, Irina-Elena Veliche, Itai Gat, Jake Weissman, James Geboski, James Kohli, Janice Lam, Japhet Asher, Jean-Baptiste Gaya, Jeff Marcus, Jeff Tang, Jennifer Chan, Jenny Zhen, Jeremy Reizenstein, Jeremy Teboul, Jessica Zhong, Jian Jin, Jingyi Yang, Joe Cummings, Jon Carvill, Jon Shepard, Jonathan McPhie, Jonathan Torres, Josh Ginsburg, Junjie Wang, Kai Wu, Kam~Hou U, Karan Saxena, Kartikay Khandelwal, Katayoun Zand, Kathy Matosich, Kaushik Veeraraghavan, Kelly Michelena, Keqian Li, Kiran Jagadeesh, Kun Huang, Kunal Chawla, Kyle Huang, Lailin Chen, Lakshya Garg, Lavender A, Leandro Silva, Lee Bell, Lei Zhang, Liangpeng Guo, Licheng Yu, Liron Moshkovich, Luca Wehrstedt, Madian Khabsa, Manav Avalani, Manish Bhatt, Martynas Mankus, Matan Hasson, Matthew Lennie, Matthias Reso, Maxim
  Groshev, Maxim Naumov, Maya Lathi, Meghan Keneally, Miao Liu, Michael~L. Seltzer, Michal Valko, Michelle Restrepo, Mihir Patel, Mik Vyatskov, Mikayel Samvelyan, Mike Clark, Mike Macey, Mike Wang, Miquel~Jubert Hermoso, Mo~Metanat, Mohammad Rastegari, Munish Bansal, Nandhini Santhanam, Natascha Parks, Natasha White, Navyata Bawa, Nayan Singhal, Nick Egebo, Nicolas Usunier, Nikhil Mehta, Nikolay~Pavlovich Laptev, Ning Dong, Norman Cheng, Oleg Chernoguz, Olivia Hart, Omkar Salpekar, Ozlem Kalinli, Parkin Kent, Parth Parekh, Paul Saab, Pavan Balaji, Pedro Rittner, Philip Bontrager, Pierre Roux, Piotr Dollar, Polina Zvyagina, Prashant Ratanchandani, Pritish Yuvraj, Qian Liang, Rachad Alao, Rachel Rodriguez, Rafi Ayub, Raghotham Murthy, Raghu Nayani, Rahul Mitra, Rangaprabhu Parthasarathy, Raymond Li, Rebekkah Hogan, Robin Battey, Rocky Wang, Russ Howes, Ruty Rinott, Sachin Mehta, Sachin Siby, Sai~Jayesh Bondu, Samyak Datta, Sara Chugh, Sara Hunt, Sargun Dhillon, Sasha Sidorov, Satadru Pan, Saurabh Mahajan,
  Saurabh Verma, Seiji Yamamoto, Sharadh Ramaswamy, Shaun Lindsay, Shaun Lindsay, Sheng Feng, Shenghao Lin, Shengxin~Cindy Zha, Shishir Patil, Shiva Shankar, Shuqiang Zhang, Shuqiang Zhang, Sinong Wang, Sneha Agarwal, Soji Sajuyigbe, Soumith Chintala, Stephanie Max, Stephen Chen, Steve Kehoe, Steve Satterfield, Sudarshan Govindaprasad, Sumit Gupta, Summer Deng, Sungmin Cho, Sunny Virk, Suraj Subramanian, Sy~Choudhury, Sydney Goldman, Tal Remez, Tamar Glaser, Tamara Best, Thilo Koehler, Thomas Robinson, Tianhe Li, Tianjun Zhang, Tim Matthews, Timothy Chou, Tzook Shaked, Varun Vontimitta, Victoria Ajayi, Victoria Montanez, Vijai Mohan, Vinay~Satish Kumar, Vishal Mangla, Vlad Ionescu, Vlad Poenaru, Vlad~Tiberiu Mihailescu, Vladimir Ivanov, Wei Li, Wenchen Wang, Wenwen Jiang, Wes Bouaziz, Will Constable, Xiaocheng Tang, Xiaojian Wu, Xiaolan Wang, Xilun Wu, Xinbo Gao, Yaniv Kleinman, Yanjun Chen, Ye~Hu, Ye~Jia, Ye~Qi, Yenda Li, Yilin Zhang, Ying Zhang, Yossi Adi, Youngjin Nam, Yu, Wang, Yu~Zhao, Yuchen Hao, Yundi
  Qian, Yunlu Li, Yuzi He, Zach Rait, Zachary DeVito, Zef Rosnbrick, Zhaoduo Wen, Zhenyu Yang, Zhiwei Zhao, and Zhiyu Ma.
\newblock The llama 3 herd of models, 2024.

\bibitem{10.1111/j.2517-6161.1996.tb02080.x}
Robert Tibshirani.
\newblock Regression shrinkage and selection via the lasso.
\newblock {\em Journal of the Royal Statistical Society: Series B (Methodological)}, 58(1):267--288, 12 2018.

\end{thebibliography}

\appendix
\section*{Appendix}

\subsection*{Supplementary Results}

We present supplementary figures corresponding to the key causal experiments.  
Each experiment isolates a distinct property of the alignment mechanism within the preference-aligned model.

\paragraph{A1: Directionality}
\label{appendix:c}
Patching activations from the aligned model into the base consistently increases $\Delta \log p$, 
whereas the reverse direction (Base$\rightarrow$Aligned) yields minimal change.  
This confirms that alignment information flows asymmetrically, from tuned mid-layer representations into the base model’s output computation.  
\begin{figure}[h]
    \centering
    \includegraphics[width=0.45\linewidth]{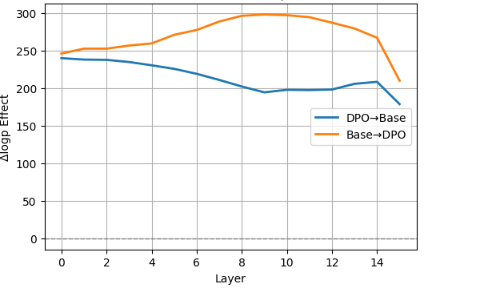}
    \caption{Patching from the aligned model into the base increases $\Delta \log p$, demonstrating unidirectional causal flow of alignment information.}
\end{figure}

\paragraph{A2: Identity and Random Controls.}
\label{appendix:a}
Control interventions demonstrate that alignment effects are not one off or isolated, at least within the context of our experiment;  
replacing activations with Gaussian noise or self-patching (Base$\rightarrow$Base) produces near-zero $\Delta \log p$.  
This validates that observed effects stem from genuine representational transfer rather than random activation perturbations.  
\begin{figure}[h]
    \centering
    \includegraphics[width=0.45\linewidth]{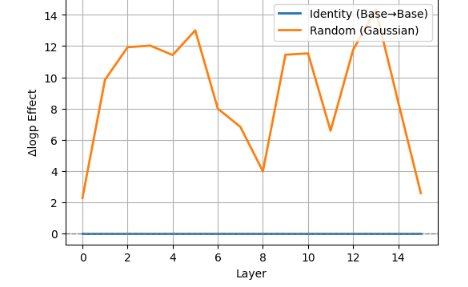}
    \caption{Identity and random control patching confirming causal specificity and non-spurious alignment transfer.}
\end{figure}

\paragraph{A3: $\boldsymbol{\alpha}$-Interpolation.}
\label{appendix:b}
To test for continuity, we interpolate between base and aligned activations with a mixing coefficient $\alpha \in [0, 1]$.  
The increase in $\Delta \log p$ with $\alpha$ demonstrates that alignment behavior scales proportionally with patch strength, 
consistent with previous results.  
\begin{figure}[h]
    \centering
    \includegraphics[width=0.45\linewidth]{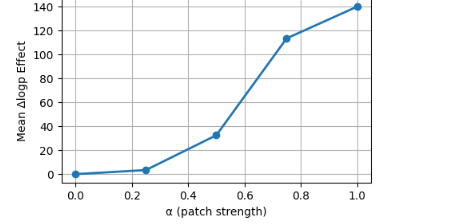}
    \caption{Dose–response relationship between patch strength $\alpha$ and alignment gain, confirming continuous dependency on representation strength.}
\end{figure}

---

\subsection*{Reproducibility Details}

All experiments were implemented in Python using the Hugging Face Transformers library.  
Causal interventions were applied with layerwise activation replacement hooks at model forward pass time.  
Statistical regression and SVD analyses were conducted using scikit-learn and NumPy, 
and all visualizations were generated via Matplotlib.

We selected the LASSO regularization parameter $\lambda$ via 5-fold cross-validation using the \texttt{LassoCV} implementation in \texttt{scikit-learn}.
The cross-validation was performed over a log-spaced grid of $\lambda$ values in the range $[10^{-4}, 10^{1}]$, following z-scoring of per-layer features $\|\Delta h_\ell\|_2$ and mean-centering of the target variable ($\Delta \log p$).
The reported coefficients correspond to the model’s cross-validation–optimal regularization parameter ($\alpha\_$ in \texttt{LassoCV}).

\subsection*{Compute and Resources}

Experiments were performed on a single NVIDIA A100 GPU (80 GB) with mixed precision (fp16) enabled.  
Each evaluation batch contained one prompt pair, and per-layer patching was applied sequentially.  

\end{document}